\documentclass[a4paper,10pt]{article}

\usepackage{ifxetex}
\usepackage{amsmath,amsfonts,amssymb,amsthm}

\ifxetex  %
   \usepackage{fontspec}
   \defaultfontfeatures{Mapping=tex--text}   %
   \usepackage{xltxtra} %
\else  %
   \usepackage[T1]{fontenc}
   \usepackage[utf8]{inputenc}
   \usepackage{lmodern} %
\fi

\usepackage[british]{babel}

\usepackage[all]{nowidow}  %
\usepackage{indentfirst}
\usepackage{url}
\usepackage{graphicx}
\usepackage{booktabs}    %
\usepackage{accents}    %
\usepackage{tikz}
\usepackage{pgfplots}

\usepackage{url}
\usepackage{hyperref}
\usepackage[square,numbers,sort&compress]{natbib}
\usepackage{subcaption}

\usepackage{thmtools}                    %

\declaretheorem[style=definition,qed=$\blacksquare$,numberwithin=section,name=Example]{example}

\newcommand{\linguistic}[1]{\texttt{#1}}

\newcommand{\KSorcid}{0000-0002-6118-606X} %

\newcommand{\zbior}[1]{\mathbb{#1}}  %

\newcommand{\fourgausses}{`4 Gausses'} 
\newcommand{\graf}{.}

\usepackage{color}
\definecolor{brickred}      {cmyk}{0   , 0.89, 0.94, 0.28}

\makeatletter \newcommand \kslistofremarks{\section*{Uwagi} \@starttoc{rks}}
  \newcommand\l@uwagas[2]
    {\par\noindent \textbf{#2:} %
{#1}\par} \makeatother

\title{\sffamily\bfseries Automatic Extraction of Linguistic Description  from Fuzzy Rule Base}
\author{Krzysztof Simiński\\
Silesian University of Technology, Poland\\
\url{krzysztof.siminski@polsl.pl}\\
orcid: \KSorcid 
\and 
Konrad Wnuk\\
Silesian University of Technology, Poland 
}
\date{\today}

\begin{document}
 
\maketitle

\begin{abstract}
Neuro-fuzzy systems are a technique of explainable artificial intelligence (XAI). They elaborate knowledge models as a set of fuzzy rules.
Fuzzy sets are crucial components of fuzzy rules. They are used to model linguistic terms. 
In this paper, we present an automatic extraction of fuzzy rules in the natural English language.
Full implementation is available free from a public repository.
\end{abstract}

\section{Introduction}

Nowadays, artificial intelligence is a very fast-developing field in computer research. 
Tools of artificial intelligence (AI) are commonly based on knowledge models. 
They may be completely unreadable for humans (eg weights of intersynaptic links in artificial neural networks) or may have a human-friendly form (eg decision trees, rules).

Neuro-fuzzy systems are a method of artificial intelligence.
They elaborate intelligible models based on fuzzy rules. 
The rules can be read and interpreted by humans. 
Thus, neuro-fuzzy systems are an example of explainable artificial intelligence (XAI).

In this paper, we present an automatic transformation of rules elaborated by NFS into linguistic sentences in the natural English language.

\section{Neuro-fuzzy systems}
\label{id:sec:nfs}
Neuro-fuzzy systems (NFS) are one of the tools of XAI.
They can handle imprecision and uncertainty of data due to fuzzy sets (the “fuzzy” part). 
They can also automatically elaborate internal representation of knowledge  from presented data (the “neuro” part). 
They elaborate data models composed of IF-THEN rules that can be interpreted by humans. 
Rules are composed of linguistic terms. 

\begin{example}
Let's analyse the fuzzy rule presented below.
\begin{center}
\textbf{if} \textit{day} \textbf{is}  \texttt{long} \textbf{and} \textit{wind} \textbf{is} \texttt{weak} \textbf{then} \textit{temperature} \textbf{is} \texttt{high}
\end{center}

The part introduced by \textbf{if} is the premise: “\textit{day} \textbf{is}  \texttt{long} \textbf{and} \textit{wind} \textbf{is} \texttt{weak}”.
The keyword \textbf{then} starts the consequence: “\textit{temperature} \textbf{is} \texttt{high}”.

The premise is composed of two atomic parts: “\textit{day} \textbf{is}  \texttt{long}” and “\textit{wind} \textbf{is} \texttt{weak}”.
Each of them has an attribute (italicised): \textit{day}, \textit{wind}%
, and a descriptor (capitalised): \texttt{long}, \texttt{weak}. %
The same applies to consequences. The only difference is that a consequence is always simple and has only one attribute.
\end{example}
Descriptors (both in premises and consequences of fuzzy rules) are not defined in a crisp numerical way. They are defined with fuzzy sets. 
This makes fuzzy and neuro-fuzzy systems robust to imprecision and uncertainty of the data they work on  \cite{id:Grzegorzewski2020Flexible, id:Sholla2020Neuro}.
Application of fuzzy sets in rules makes them easier to interpret for humans \cite{id:Siminski2021Outlier}. 
It is a kind of interface between humans and computers. 
Humans can easily write down their knowledge in the form of fuzzy rules that can be incorporated into the fuzzy rule base of a (neuro-)fuzzy system.
And the other way round: Rules elaborated automatically by NFS can be transformed into a linguistic description.   
The values in fuzzy rules are not purely numerical abstract parameters. 
These values have significant meanings. 
Fuzzy rules can be read in a natural human language.
This makes a fuzzy rule base interpretable \cite{id:Cpalka2014New,id:Alcala2006Hybrid,id:Gacto2011Interpretability,id:Alonso2015Interpretability,id:Magdalena2021Fuzzy,id:Lapa2018New,id:Riid2002Transparent,id:Slowik2020Multi,id:Leski2015Fuzzy,id:Alonso2011Special,id:Siminski2022FuBiNFS,id:Evsukoff2009designofinterpretable,id:Bartczuk2016New}.
Neuro-fuzzy systems are an example of explainable artificial intelligence (XAI).

\section{Linguistic description of fuzzy rules}
Descriptors are implemented with fuzzy sets. 
In the premises of rules, any fuzzy set can be used. 
The choice depends on many factors (eg some NFS use gradient optimisation techniques that require the membership functions of fuzzy sets to be differentiable).  
In consequences, fuzzy sets are limited to triangular fuzzy sets (Mamdami-Assilan NFS) and fuzzy singletons (Takagi-Sugeno-Kang NFS). The reason is to avoid numerical integration and to keep computation complexity low. 

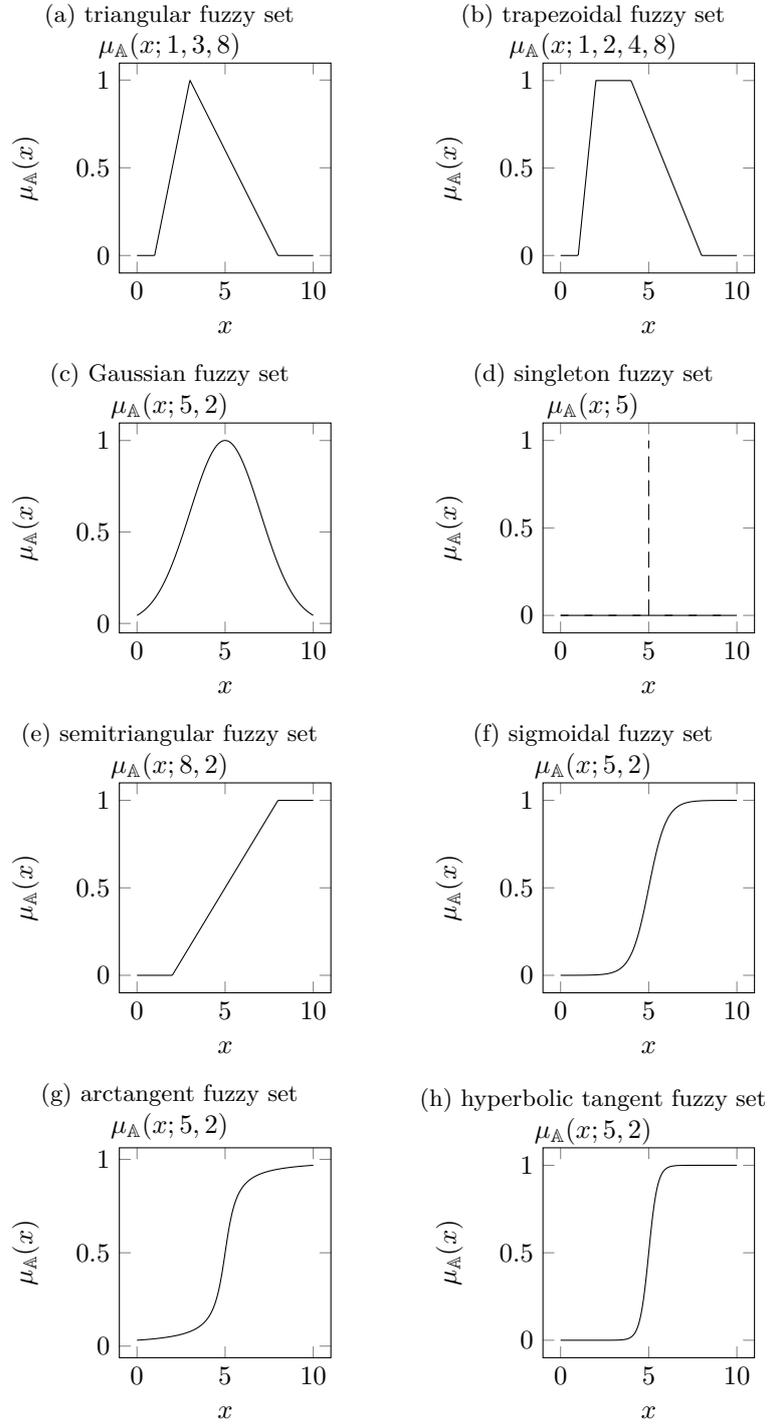
\begin{figure}
\centering
\begin{subfigure}{0.45\textwidth}
     \begin{center}
    \caption{triangular fuzzy set}
    \label{id:fig:triangular}
		$\mu_\zbior{A}(x; 1,3,8)$
		\begin{tikzpicture}
		\begin{axis}[width=0.8\textwidth,height=0.8\textwidth,xlabel=$x$,ylabel={$\mu_\zbior{A}(x)$}]
		    \addplot[domain=0:10,samples=200] {max (min((x-1)/2, (8-x)/5), 0)};
	\end{axis}
	\end{tikzpicture}
	\end{center}
\end{subfigure}
\begin{subfigure}{0.45\textwidth}
     \begin{center}
    \caption{trapezoidal fuzzy set}
    \label{id:fig:trapezoidal}
		$\mu_\zbior{A}(x; 1,2,4,8)$
		\begin{tikzpicture}
		\begin{axis}[width=0.8\textwidth,height=0.8\textwidth,xlabel=$x$,ylabel={$\mu_\zbior{A}(x)$}]
    \addplot[domain=0:10,samples=200] {max (min((x-1), 1,  (8-x)/4), 0)};
\end{axis}
	\end{tikzpicture}
	\end{center}
\end{subfigure}

\begin{subfigure}{0.45\textwidth}
     \begin{center}
    \caption{Gaussian fuzzy set}
    \label{id:fig:gauss}
		$\mu_\zbior{A}(x; 5,2)$
		\begin{tikzpicture}
		\begin{axis}[width=0.8\textwidth,height=0.8\textwidth,xlabel=$x$,ylabel={$\mu_\zbior{A}(x)$}]
		   \addplot[domain=0:10,samples=200] {exp (-((x-5)^2)/(8)};
	\end{axis}
	\end{tikzpicture}
	\end{center}
\end{subfigure}
\begin{subfigure}{0.45\textwidth}
     \begin{center}
    \caption{singleton fuzzy set}
    \label{id:fig:singleton}
		$\mu_\zbior{A}(x; 5)$
		\begin{tikzpicture}
		\begin{axis}[width=0.8\textwidth,height=0.8\textwidth,xlabel=$x$,ylabel={$\mu_\zbior{A}(x)$}]
        \addplot[mark=none,loosely dashed] plot coordinates {(0,0) (5,0) (5, 1) (5,0) (10,0)};
        \addplot[mark=none] plot coordinates {(0,0) (10,0)};
    	\end{axis}
	\end{tikzpicture}
	\end{center}
\end{subfigure}

\begin{subfigure}{0.45\textwidth}
     \begin{center}
    \caption{semitriangular fuzzy set}
    \label{id:fig:semitriangular}
		$\mu_\zbior{A}(x; 8,2)$
		\begin{tikzpicture}
		\begin{axis}[width=0.8\textwidth,height=0.8\textwidth,xlabel=$x$,ylabel={$\mu_\zbior{A}(x)$}]
    \addplot[domain=0:10,samples=200] {max (min(1,  (2-x)/(-6)), 0)};
\end{axis}
	\end{tikzpicture}
	\end{center}
\end{subfigure}
\begin{subfigure}{0.45\textwidth}
     \begin{center}
    \caption{sigmoidal fuzzy set}
    \label{id:fig:sigmoidal}
		$\mu_\zbior{A}(x; 5,2)$
		\begin{tikzpicture}
		\begin{axis}[width=0.8\textwidth,height=0.8\textwidth,xlabel=$x$,ylabel={$\mu_\zbior{A}(x)$}]
    \addplot[domain=0:10,samples=200] {1 / (1 + exp(-2 * (x - 5)))};
\end{axis}
	\end{tikzpicture}
	\end{center}
\end{subfigure}

\begin{subfigure}{0.45\textwidth}
     \begin{center}
    \caption{arctangent fuzzy set}
    \label{id:fig:arctan}
		$\mu_\zbior{A}(x; 5,2)$
		\begin{tikzpicture}
		\begin{axis}[width=0.8\textwidth,height=0.8\textwidth,xlabel=$x$,ylabel={$\mu_\zbior{A}(x)$}]
    \addplot[domain=0:10,samples=200] {0.5 + (rad(atan(2*(x-5))))/3.14};
\end{axis}
	\end{tikzpicture}
	\end{center}
\end{subfigure}
\begin{subfigure}{0.45\textwidth}
     \begin{center}
    \caption{hyperbolic tangent fuzzy set}
    \label{id:fig:tanh}
		$\mu_\zbior{A}(x; 5,2)$
		\begin{tikzpicture}
		\begin{axis}[width=0.8\textwidth,height=0.8\textwidth,xlabel=$x$,ylabel={$\mu_\zbior{A}(x)$}]
     \addplot[domain=0:10,samples=200] {0.5 + 0.5 *(tanh(2*(x-5)))};
    \end{axis}
	\end{tikzpicture}
	\end{center}
\end{subfigure}

\caption{Examples of membership functions in fuzzy descriptors.}
\label{id:fig:membership-functions}
\end{figure}

Tab. \ref{id:tab:membership-functions} presents maths formulae for the fuzzy sets and Fig. \ref{id:fig:membership-functions} – plots.
Triangular (Fig. \ref{id:fig:triangular}), 
trapezoidal (Fig. \ref{id:fig:trapezoidal}), 
Gaussian (Fig. \ref{id:fig:gauss}), and 
singleton  (Fig. \ref{id:fig:singleton}) fuzzy sets are used to model two-tailed limited linguistic notion like eg \texttt{low} or \texttt{high}.
Semitriangular (Fig. \ref{id:fig:semitriangular}),
sigmoidal (Fig. \ref{id:fig:singleton}),
arctangent (Fig. \ref{id:fig:arctan}), and
hyperbolic tangent (Fig. \ref{id:fig:tanh})
are used to model one-tailed limited linguistic notions like eg \texttt{less than} or \texttt{greater than}.
Gaussian, sigmoidal, arctangent, and hyperbolic tangent membership functions are differentiable in their whole domain.

\begin{table}
\centering
\caption{Examples of membership functions in fuzzy descriptors.}
\label{id:tab:membership-functions}
\begin{tabular}{lll}
	\toprule
	name        & formula & figure \\
	\midrule
	triangular  & $\mu_\zbior{A} (x; a,b,c) = \max \left[ \min \left( \frac{x -a}{b-a}, \frac{c- x}{c - b} \right), 0 \right] $        &  Fig. \ref{id:fig:triangular}       \\
	trapezoidal & $\mu_\zbior{A} (x; a,b,c,d) =\max \left[ \min \left( \frac{x -a}{b-a}, 1, \frac{d- x}{d - c} \right), 0 \right]$        &    Fig. \ref{id:fig:trapezoidal}    \\
	Gaussian    &  $\mu_\zbior{A} (x; m, \sigma) =\exp \left( - \frac{(x - m)^2}{2\sigma^2} \right) $  & Fig. \ref{id:fig:gauss}     \\
	singleton   & $ \mu_\zbior{A} (x; a) = \begin{cases}1, & x = a\\	0, & x \neq a	\end{cases} $  & Fig. \ref{id:fig:singleton} \\
\midrule
	semitriangular & $\mu_\zbior{A} (x; a,b) =\max \left[ \min \left(  1, \frac{b- x}{b - a} \right), 0 \right]$ & Fig. \ref{id:fig:semitriangular} \\
	sigmoidal   & $\mu_\zbior{A} (x; c, s) = \frac{1}{1 + \exp(-s(x-c))}$ & Fig. \ref{id:fig:sigmoidal}\\
    arctangent & $\mu_\zbior{A} (x; c, s) = 0.5 + \frac{1}{\pi} \arctan \left( s \left(x - c\right) \right)$ & Fig. \ref{id:fig:arctan} \\
	hyperbolic tangent & $\mu_\zbior{A}(x; c, s) = \frac{1}{2} + \frac{1}{2} \tanh \left( s (x - c) \right) $ & Fig. \ref{id:fig:tanh} \\
	\bottomrule
\end{tabular}
\end{table}

The elaboration of linguistic descriptions of descriptors starts with the calculation of the average and standard deviation for each attribute in the dataset.
The localisation of the fuzzy descriptor is labelled as
\linguistic{micro},
\linguistic{tiny},
\linguistic{small},
\linguistic{medium},
\linguistic{large},
\linguistic{huge}, and 
\linguistic{giant}.
 
\subsection{Two-tailed descriptors}
A two-tailed descriptor has asymptotic values 
$\lim_{x \to \infty} = \lim_{x \to -\infty} = 0$. 

The fuzziness of the descriptors is described with labels:
\linguistic{strictly},  %
\linguistic{distinctly},  %
\linguistic{moderately},  %
\linguistic{mildly}, and %
\linguistic{loosely}.  %

\subsubsection{Fuzzy singleton}
The linguistic description is very simple for a fuzzy singleton descriptor. A singleton is described with \linguistic{is exactly}.

\subsubsection{Gaussian descriptor}

The location label $l_m$ is elaborated with the formula:
\begin{align}
l_c = 2\frac{\bar{x} - m}{\hat{x}} + \left\lfloor \frac{7}{2} \right\rfloor,
\end{align}
where $\bar{x}$ is the average and $\hat{x}$ is the standard deviations of the attribute $x$ and 7 is the number of labels in the description set; $m$ stands for the core of the Gaussian fuzzy set.

The fuzziness is directly mapped from the standard deviation $\hat{x}$ of the attribute in question (Tab. \ref{id:tab:gaussian:fuzziness}).

\begin{table}
\centering
\caption{Fuzziness mapping for the Gaussian descriptor.}
\label{id:tab:gaussian:fuzziness}
\begin{tabular}{cl}
\toprule 
$\hat{x}$ & label \\
\midrule
$(-\infty, 0.5)$ & \linguistic{strictly} \\   %
$[0.5, 1)$ &  \linguistic{distinctly} \\ %
$[1, 2)$ & \linguistic{moderately} \\ %
$[2, 5)$ & \linguistic{mildly}     \\ %
$[5, \infty)$ & \linguistic{loosely}    \\ %
\bottomrule
\end{tabular}
\end{table}

\subsubsection{Triangular descriptor}
The triangular descriptor is similar to the Gaussian one but it is not necessarily symmetrical. The location $l$ is elaborated as the centre of gravity of a  triangular area:
\begin{align}
l = (s_{\min} + c + s_{\max}) / 3,
\end{align}
where $s_{\min}$ and $s_{\max}$ stand for the minimum and maximum of the support, $c$ stands for the core of the triangular fuzzy  set.

This descriptor is also used for triangular fuzzy sets in consequences of a Mamdami-Assilan (neuro-)fuzzy system.

\subsubsection{Trapezoidal descriptor}
This descriptor is similar to the triangular one. The location $l$ is elaborated with the formula
\begin{align}
l = (s_{\min} + c_{\min} + c_{\min} + s_{\max}) / 4,
\end{align}
where $s_{\min}$ and $s_{\max}$ stand for the minimum and maximum of the support, $c_{\min}$ and $c_{\min}$ stand for the minimum and maximum of  the core of the trapezoidal fuzzy  set.

\subsection{One-tailed descriptors}
A one-tailed descriptor $d$ has asymptotic values 
$\lim_{x \to \infty} d(x) = 0 \land \lim_{x \to -\infty} d(x) = 1$ 
or 
$\lim_{x \to \infty} d(x) = 1 \land \lim_{x \to -\infty} d(x) = 0$. 
It has also one value (crosspoint) $c$ such as $ d(c) = 0.5$ (Tab. \ref{id:tab:membership-functions}). 
For these descriptors the location of the crosspoint is described with linguistic terms: 
\linguistic{micro},
\linguistic{tiny},
\linguistic{small},
\linguistic{medium},
\linguistic{large},
\linguistic{huge}, and 
\linguistic{giant}.
The slope ($s$, Tab. \ref{id:tab:membership-functions}) is described with linguistic terms:
\linguistic{hardly},       
\linguistic{mildly},       
\linguistic{moderately},   
\linguistic{distinctly}, and    
\linguistic{stepwise}.

\subsubsection{Sigmoidal, semitriangular, and arctangent descriptors}
The location label $l_c$ is elaborated with the formula:
\begin{align}
l_c = \frac{\bar{x} - c}{\bar{x}} + \left\lfloor \frac{7}{2} \right\rfloor,
\end{align}
where $\bar{x}$ is the average value of the attribute $x$ and 7 is the number of labels in the description set.

The slope label $l_s$ is calculated as
\begin{align}
l_s = |s \cdot \hat{x}|,
\end{align}
where $\hat{x}$ is the standard deviation of the attribute $x$.
Then the $l_s$ is mapped as presented in Tab. \ref{id:fig:sigm:slope-mapping}.

\begin{table}
\centering
\caption{Slope mapping for the sigmoidal, arctangent, and hyperbolic tangent descriptors.}
\label{id:fig:sigm:slope-mapping}
\begin{tabular}{ccl}
	\toprule
	        \multicolumn{2}{c}{$l_c$}          & label                   \\
	sigmoidal, semitriangular, and $\arctan$ &     $\tanh$      &                         \\
	\midrule
	    $[10, \infty)$      &  $[5, \infty)$   & \linguistic{hardly}     \\
	       $[4, 10)$        &     $[2, 5)$     & \linguistic{mildly}     \\
	       $[1, 4)$         &    $[0.5, 2)$    & \linguistic{moderately} \\
	      $[0.4, 1)$        &   $[0.2, 0.5)$   & \linguistic{distinctly} \\
	   $(-\infty, 0.4)$     & $(-\infty, 0.2)$ & \linguistic{stepwise}   \\
	\bottomrule
\end{tabular}
\end{table} 

\subsubsection{Hyperbolic tangent descriptor}
The elaboration is the same as in the sigmodal descriptor with the only 
diffrence being the mapping as stated in Tab. \ref{id:fig:sigm:slope-mapping}.

\begin{figure}
\centering 
 \includegraphics[width=\textwidth]{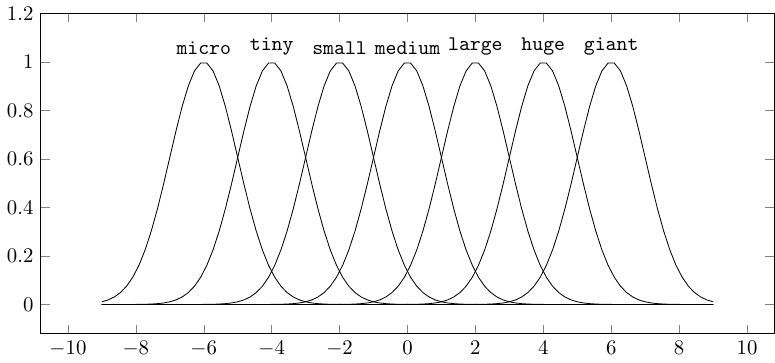}
\caption{The semantics of linguistic labels for location applied to a Gaussian descriptor.} 
\label{id:fig:labels:location:gaussian}
\end{figure}

\begin{figure}
\centering 
\includegraphics{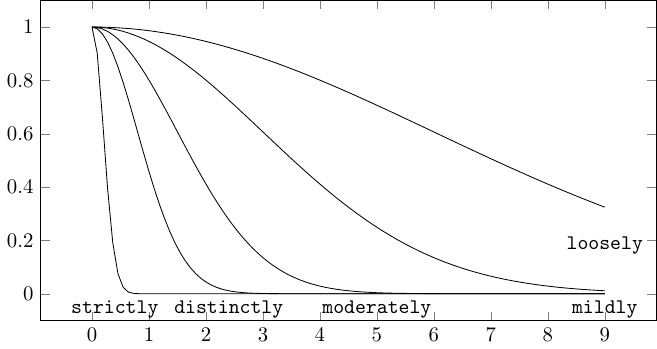}
\caption{The semantics of linguistic labels for fuzziness applied to a (half of a) Gaussian descriptor.} 
\label{id:fig:labels:fuzziness:gaussian}
\end{figure}

\section{Example}

The implementation of neuro-fuzzy systems used in this paper is freely available from the GitHub repository\footnote{\url{https://github.com/ksiminski/neuro-fuzzy-library}}.

\subsection{Dataset}

\begin{figure}
\centering
\includegraphics[width=0.95\textwidth]{\graf/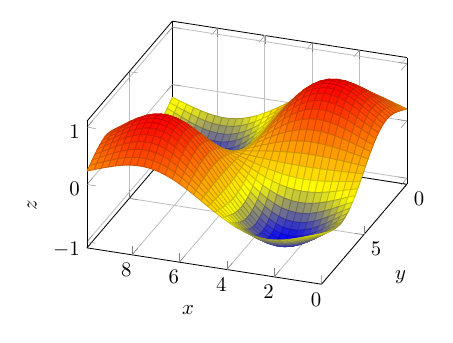}
\caption{The surface of the \fourgausses\ dataset.} 
\label{id:fig:four-gausses}
\end{figure}

The \fourgausses\ is is a synthetic dataset representing a surface  composed of four Gaussian functions:
\begin{align}
z(x,y) = g_1(x,y) + g_2(x,y) - g_3(x,y) - g_4(x,y),
\end{align}
where 
\begin{align}
g_1(x,y) = &
\exp \left( - \frac{\left(x - \frac{m}{4}\right)^2}{2\sigma^2} \right)
\cdot 
\exp \left( - \frac{\left(y - \frac{m}{4}\right)^2}{2\sigma^2} \right) 
,\\
g_2(x,y) = &
\exp \left( - \frac{\left(x - \frac{3m}{4}\right)^2}{2\sigma^2} \right)
\cdot 
\exp \left( - \frac{\left(y - \frac{3m}{4}\right)^2}{2\sigma^2} \right) 
,\\
g_3(x,y) = &
\exp \left( - \frac{\left(x - \frac{m}{4}\right)^2}{2\sigma^2} \right)
\cdot 
\exp \left( - \frac{\left(y - \frac{3m}{4}\right)^2}{2\sigma^2} \right) 
,\\
g_4(x,y) = &
\exp \left( - \frac{\left(x - \frac{3m}{4}\right)^2}{2\sigma^2} \right)
\cdot 
\exp \left( - \frac{\left(y - \frac{m}{4}\right)^2}{2\sigma^2} \right) 
,
\end{align}
where $m = 10$ and $\sigma = 2$. The surface is presented in Fig. \ref{id:fig:four-gausses}.

\subsection{Fuzzy systems}
This section provides two examples of fuzzy rule bases created manually with triangular (Sec. \ref{id:sec:triangular}) and sigmoidal (Sec. \ref{id:sec:sigmoidal}) descriptors in premises of rules. Both systems have triangular fuzzy sets in the consequences of rules.
\subsubsection{Fuzzy system with triangular descriptors in premises}
\label{id:sec:triangular}
\begin{verbatim}
RULE 1
IF     input 1 is loosely tiny 
   AND input 2 is loosely tiny 
THEN output is strictly giant.

RULE 2
IF     input 1 is loosely tiny 
   AND input 2 is loosely large 
THEN output is strictly micro.

RULE 3
IF     input 1 is loosely large 
   AND input 2 is loosely tiny 
THEN output is strictly micro.

RULE 4
IF     input 1 is loosely large 
   AND input 2 is loosely large 
THEN output is strictly giant.
\end{verbatim}

\subsubsection{Fuzzy system with sigmoidal descriptors in premises}
\label{id:sec:sigmoidal}
\begin{verbatim}
RULE 1
IF     input 1 is mildly less than medium 
   AND input 2 is mildly less than medium 
THEN output is strictly giant.

RULE 2
IF     input 1 is mildly less than medium 
   AND input 2 is mildly greater than medium 
THEN output is strictly micro.

RULE 3
IF     input 1 is mildly greater than medium 
   AND input 2 is mildly less than medium 
THEN output is strictly micro.

RULE 4
IF     input 1 is mildly greater than medium 
   AND input 2 is mildly greater than medium 
THEN output is strictly giant.
\end{verbatim}
\subsection{Neuro-fuzzy systems}
\subsubsection{Mamdani-Assilan neuro-fuzzy system}
Mamdani-Assilan NFS (MA) \cite{id:Mamdani1974Application,id:Mamdani1975experimentinlinguistic} is a system with Gaussian descriptors in the premises of rules and triangular fuzzy sets in the consequences. 
The fuzzy rule base is elaborated automatically.
The parameters are tuned with the gradient method. This type of NFS has lower precision, but higher interpretability.

The fuzzy rule base elaborated for the \fourgausses\ dataset by the Mamdani-Assilan neuro-fuzzy system is:

\begin{verbatim}

RULE 1
IF     input 1 is moderately tiny 
   AND input 2 is moderately large 
THEN output is strictly micro.

RULE 2
IF     input 1 is moderately large 
   AND input 2 is moderately large 
THEN output is strictly large.

RULE 3
IF     input 1 is moderately tiny 
   AND input 2 is moderately tiny 
THEN output is strictly large.

RULE 4
IF     input 1 is moderately large 
   AND input 2 is moderately tiny 
THEN output is strictly tiny.
\end{verbatim}

\subsubsection{Takagi-Sugeno-Kang neuro-fuzzy system}
Takagi-Sugeno-Kang NFS (TSK) is a neuro-fuzzy system with Gaussian descriptors in the premises of rules and moving singletons in the consequences. Localisations of fuzzy singletons are not constant – they depend on input values. This makes the elaborated models less interpretable, but more precise (resulting in lower numerical errors).
The fuzzy rule base is elaborated automatically.
The parameters are tuned with the gradient method for premises and the least square method for consequences \cite{id:Takagi1985fuzzyidentificationof,id:Sugeno1988structureidentificationof}.

The fuzzy rule base elaborated for the \fourgausses\ dataset  by the Takagi-Sugeno-Kang neuro-fuzzy system is:
\begin{verbatim}
RULE 1
IF     input 1 is moderately large 
   AND input 2 is moderately tiny 
THEN   input 1 has low positive importance 
   AND input 2 has low negative importance 
   AND constant term is tiny.

RULE 2
IF     input 1 is moderately tiny 
   AND input 2 is moderately large 
THEN   input 1 has low negative importance 
   AND input 2 has low positive importance 
   AND constant term is tiny.

RULE 3
IF     input 1 is moderately tiny 
   AND input 2 is moderately tiny 
THEN   input 1 has low positive importance 
   AND input 2 has low positive importance 
   AND constant term is medium.

RULE 4
IF     input 1 is moderately large 
   AND input 2 is moderately large 
THEN   input 1 has low positive importance 
   AND input 2 has low negative importance 
   AND constant term is large.
\end{verbatim}

\subsubsection{Neuro-fuzzy system with logical implications}
ANNBFIS (artificial neural network based fuzzy inference system) is a neuro-fuzzy system with Gaussian descriptors in the premises of rules and moving isosceles triangular sets in the consequences. The parameters are tuned with the gradient method for the premises and least square method for consequences \cite{id:Czogala2000Fuzzy}.
The interpretability of the model is similar to the TSK, so we do not provide an example, but the linguistic description for this system is implemented in the repository.
 
\section{Summary}
\label{id:sec:summary}
The paper presents a module for automatic extraction of linguistic description of fuzzy rule base elaborated by MA, TSK, and ANNBFIS neuro-fuzzy systems. 
The implementation is available from the GitHub repository\footnote{\url{https://github.com/ksiminski/neuro-fuzzy-library}}.
 
\bibliographystyle{plain}
\bibliography{bibliografia}
\end{document}